%% file: main.tex
\documentclass{article}

\usepackage[preprint]{neurips_2026}

\usepackage{amsmath,amssymb,amsfonts}
\usepackage{graphicx}
\usepackage{textcomp}
\usepackage{xcolor}
\usepackage[hyphens]{url}
\usepackage{booktabs}
\usepackage[colorlinks=true, urlcolor=blue, citecolor=black, linkcolor=black]{hyperref}
\usepackage{makecell}
\usepackage{multirow}
\usepackage{subcaption}
\usepackage{float}
\usepackage{nicefrac}
\usepackage{listings}
\usepackage{xspace}
\usepackage{adjustbox}
\usepackage{pifont}
\usepackage[utf8]{inputenc}
\usepackage[T1]{fontenc}
\usepackage{microtype}
\usepackage{enumitem}
\usepackage{tikz}
\usetikzlibrary{positioning,shapes.geometric,fit,calc,arrows.meta,backgrounds}
\usepackage{amsthm}
\usepackage{algorithm}
\usepackage{algorithmic}
\usepackage{cleveref}

\newtheorem{definition}{Definition}

\newcommand{\solar}{\textsc{Solar}\xspace}
\newcommand{\orojenesis}{\textsc{Orojenesis}\xspace}
\newcommand{\solbench}{\textsc{SOL-ExecBench}\xspace}

\definecolor{okblue}{HTML}{0072B2}
\definecolor{okorange}{HTML}{E69F00}
\definecolor{okgreen}{HTML}{009E73}
\definecolor{okred}{HTML}{D55E00}
\definecolor{okpurple}{HTML}{CC79A7}
\definecolor{okcyan}{HTML}{56B4E9}

\newif\ifdraftcomments
\draftcommentstrue

\title{\solar: Speed-of-Light Performance Analysis\\for Deep Learning Workloads from Source Code}

\title{\solar: AI-Powered Speed-of-Light Performance Analysis}
\begin{document}

\author{
\textbf{Qijing Huang, Sana Damani, Zhifan Ye, Athinagoras Skiadopoulos} \\ 
\textbf{Siva Kumar Sastry Hari, Jason Clemons, Sahil Modi, Jingquan Wang} \\
\textbf{Aditya Kane, Edward C Lin, Humphrey Shi, Christos Kozyrakis} \\
\\
NVIDIA
}
\maketitle
\input{sections/0_abstract}

\input{sections/1_intro}

\input{sections/2_related}

\input{sections/3_method}
\input{sections/4_eval}

\input{sections/5_conclusion}

\bibliographystyle{plainnat}
\bibliography{refs}

\newpage
\appendix       
\input{appendix}

\clearpage

\end{document}

%% file: sections/0_abstract.tex
\begin{abstract}

How fast \emph{could} a deep-learning model run on target hardware, and how far is today’s implementation from that limit? These questions are central to software, hardware, and algorithm optimizations. Speed-of-Light (SOL) analysis answers them by computing a workload’s theoretical minimum execution time on a given architecture. Yet deriving SOL bounds remains manual, error-prone, and disconnected from rapid model development.
To close this gap, we introduce SOLAR, a framework that automatically derives validated SOL bounds from Pytorch and JAX source code. 
SOLAR leverages both \emph{generative} and \emph{deterministic} components in its flow: an LLM frontend translates any source programs into an executable Affine Loop IR, validated by output comparison; a deterministic flow lifts the IR into an einsum graph; and an analytical backend computes unfused, fused, and cache-aware SOL bounds. 
SOLAR provides comprehensive operator and language coverage, produces validated bounds with zero observed SOL violations, and offers multi-fidelity analysis that tightens bounds and surfaces optimization insights.  We evaluate SOLAR across KernelBench, JAX/Flax models, and robotics workloads. These experiments demonstrate four use cases: headroom analysis at multiple fidelity levels, identifying optimization opportunities, cross-platform exploration, and inverse-roofline hardware provisioning. SOLAR is open source and available at  \url{https://github.com/NVlabs/SOLAR}.
 
\end{abstract}

%% file: sections/1_intro.tex
\section{Introduction}
\label{sec:intro}

\begin{figure}[t]
\centering
\begin{subfigure}[b]{0.38\textwidth}
    \centering
    \includegraphics[width=\textwidth]{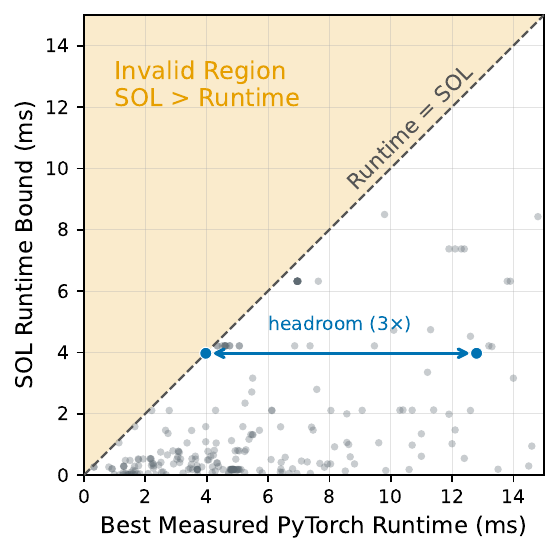}
    \caption{SOL headroom on KernelBench.}
    \label{fig:intro_scatter}
\end{subfigure}
\hfill
\begin{subfigure}[b]{0.38\textwidth}
    \centering
    \includegraphics[width=\textwidth]{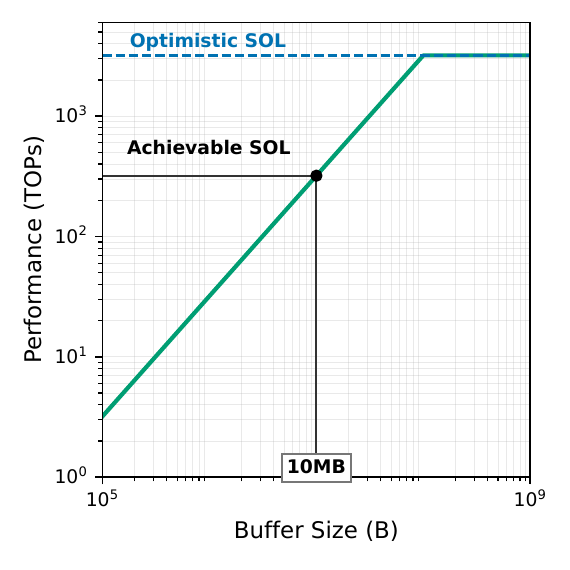}
    \caption{Na\"ive vs.\ cache-aware SOL.}
    \label{fig:orojenesis_roofline}
\end{subfigure}
\hfill
\begin{subfigure}[b]{0.19\textwidth}
    \centering
    \scriptsize
    \begin{tabular}{@{}lcc@{}}
        \toprule
        \textbf{Tool} & \textbf{Cov.} & \textbf{SOL} \\
        \midrule
        fvcore & 75\% & \texttimes \\
        ptflops & 84\% & \texttimes \\
        \solar & 100\% & \checkmark \\
        \bottomrule
    \end{tabular}
    \caption{Op coverage.}
    \label{fig:intro_table}
\end{subfigure}

\caption{\textbf{What \solar provides.}  \solar derives validated SOL bounds from source code, enabling three capabilities existing tools lack.  (a)~\emph{SOL headroom analysis}: points below the diagonal reveal optimization opportunity; \solar exposes up to orders-of-magnitude headroom on KernelBench.  (b)~\emph{Tighter bounds}: cache-aware analysis (\orojenesis) accounts for on-chip buffer constraints, tightening SOL by up to $10\times$ over na\"ive roofline.  (c)~\emph{Higher coverage}: FLOP counters~\cite{fvcore,ptflops} cover 75--84\% of KernelBench and cannot derive SOL; \solar covers 100\% with full SOL analysis.}
\label{fig:intro_combined}
\end{figure}

Modern deep-learning accelerators achieve peak throughputs measured in petaFLOPS, yet real workloads rarely approach these limits. How fast \emph{could} a model or kernel run on target hardware, and how far is today’s implementation from that limit? \emph{Speed-of-Light} (SOL) analysis answers these questions by computing a workload’s theoretical minimum execution time on a given architecture. These bounds are useful across the optimization stack: kernel and compiler engineers use them to identify bottlenecks and remaining headroom; 
algorithm designers use them to evaluate accuracy-performance-cost trade-offs;
architects use them to provision compute and bandwidth for target workloads;  and AI agents can use them to validate that generated optimizations remain physically attainable. Without analytical ceilings, optimization---and increasingly, agentic code generation---risks burning compute and tokens on work with little remaining headroom~\cite{ucutlass}.

Despite its importance, no existing tool derives SOL bounds automatically from source code.
FLOP counters~\cite{fvcore,ptflops} suffer from narrow language and operator coverage (\Cref{fig:intro_table}) and ignore I/O traffic; profilers~\cite{ncu2023} measure achieved rather than theoretical performance; and pure LLM estimation incurs errors.
Meanwhile, the growing diversity of AI architectures, including Transformers~\cite{vaswani2017attention}, MoE~\cite{deepseekv3}, SSMs~\cite{mamba2}, and hybrid models~\cite{nemotronh}, along with emerging domains such as robotics models, makes automated SOL analysis increasingly critical.  



To address this gap, we introduce \solar, \textbf{S}peed-of-Light \textbf{A}nalysis for \textbf{R}untime, a source-to-SOL framework with three goals: 1) broad operator and language coverage, 2) automated and validated SOL derivation, 3) and tighter bounds that surface optimization insights. \solar achieves these goals by separating \emph{generative} translation with validation from \emph{deterministic} analysis: 
an LLM translates source code into an executable Affine Loop IR, where correctness can be verified by numerical output comparison.
a deterministic compiler then lifts the validated IR into an Einsum graph, and an analytical backend computes unfused, fused, and cache-aware SOL    
bounds.   

This three-stage structure captures the best of both worlds: an agentic flow with output validation for easy translation from source languages to an 
Einsum representation, paired with deterministic lifting and performance analysis for reproducibility and correctness guarantees.                       
\Cref{fig:intro_scatter} illustrates the result: each point plots a KernelBench workload's measured PyTorch runtime against \solar's fused SOL bound,
with the gap to the diagonal representing optimization headroom---up to orders of magnitude on real workloads.                                      
\Cref{fig:orojenesis_roofline} further shows that cache-aware analysis \cite{orojenesis} in SOLAR tightens these bounds over na\"ive roofline
by accounting for finite on-chip buffer capacity. This paper makes three contributions:
\begin{enumerate}[leftmargin=*,itemsep=2pt,topsep=2pt]
    \item \textbf{A source-to-SOL framework.}
    \solar is the first tool to derive validated SOL bounds directly from PyTorch and JAX source code.
    It supports operator- and graph-level analysis with unfused, fused, and cache-aware bounds, achieving 100\% analysis coverage on KernelBench. 

    \item \textbf{A three-stage pipeline separating generative and deterministic reasoning.}
    Rather than asking an LLM to estimate SOL directly, \solar confines the LLM to a verifiable task: translating source code into an executable Affine Loop IR (\emph{Validated Agentic Lowering}).
    A deterministic compiler then lifts the IR into an Einsum graph (\emph{Deterministic Einsum Lifting}), from which an analytical backend derives multi-fidelity SOL bounds.

    \item \textbf{Evidence across kernel, algorithm, and architecture use cases.}
    Fusion analysis reveals up to $7.8\times$ additional headroom beyond per-operator analysis on KernelBench; cache-aware bounds tighten SOL by $2.06\times$.
    Einsum chain reordering identifies a $2.04\times$ FLOP reduction on DeepSeek MLA.
    Cross-platform projection spans four hardware targets without physical access, and inverse roofline reveals that 500\,Hz robotics VLA targets require up to $19.7\times$ current bandwidth.
\end{enumerate}

%% file: sections/2_related.tex
\section{Related Work}
\label{sec:related}

No existing tool derives validated SOL bounds directly from source code.
\Cref{tab:landscape} positions \solar against the current landscape.

\begin{table}[t]
    \centering
    \caption{\textbf{Performance analysis landscape.}  \solar is the only approach that combines execution-validated translation, graph-level einsum extraction, multi-fidelity SOL bounds, cache-aware tightening, and language-agnostic input.  ``Graph'' = analyzes operator graphs (not just individual kernels); ``SOL'' = derives theoretical bounds (not achieved performance); ``Validated'' = execution-validated translation; ``Cache'' = tighter SOL via buffer-size-aware tiling.}
    \label{tab:landscape}
    \scriptsize
    \begin{tabular}{@{}llccccc@{}}
        \toprule
        \textbf{Category} & \textbf{Tool} & \textbf{Graph} & \textbf{SOL} & \textbf{Valid.} & \textbf{Cache} & \textbf{Input} \\
        \midrule
        \multirow{3}{*}{FLOP counters} & fvcore~\cite{fvcore} & \texttimes & \texttimes & N/A & \texttimes & PyTorch \\
        & ptflops~\cite{ptflops} & \texttimes & \texttimes & N/A & \texttimes & PyTorch \\
        & torchinfo~\cite{torchinfo} & \texttimes & \texttimes & N/A & \texttimes & PyTorch \\
        \midrule
        \multirow{2}{*}{Profilers} & NCU~\cite{ncu2023} & \texttimes & \texttimes & N/A & \texttimes & Any \\
        & NSys~\cite{nsys2023} & \checkmark & \texttimes & N/A & \texttimes & Any \\
        \midrule
        \multirow{2}{*}{Predictors} & Paleo~\cite{qi2017paleo} & \checkmark & \texttimes & \texttimes & \texttimes & Caffe \\
        & Habitat~\cite{yu2021habitat} & \checkmark & \texttimes & \texttimes & \texttimes & PyTorch \\
        \midrule
        \multirow{3}{*}{LLM-based} & Pure LLM SOL & \texttimes & \checkmark & \texttimes & \texttimes & Any \\
        & ASAP~\cite{asap2025} & \checkmark & \texttimes & \texttimes & \texttimes & JAX/XLA \\
        & Opal~\cite{opal2025} & \texttimes & \texttimes & \texttimes & \texttimes & CUDA/HIP \\
        \midrule
        \multirow{2}{*}{\makecell[l]{Perf.\ modeling\\frameworks}} & Timeloop~\cite{parashar2019timeloop} & \texttimes & \checkmark & \checkmark & \checkmark & Manual \\
        & AccelForge~\cite{accelforge} & \checkmark & \checkmark & \checkmark & \checkmark & Manual \\
        \midrule
        & \textbf{\solar} & \checkmark & \checkmark & \checkmark & \checkmark & \textbf{Any} \\
        \bottomrule
    \end{tabular}
\end{table}

\begin{table}[t]
\centering
\begin{minipage}[t]{0.52\textwidth}
    \centering
    \caption{\textbf{FLOP counter coverage on KernelBench.}  fvcore and ptflops return 0 for element-wise ops and crash on L4.  Neither reports I/O traffic or graph structure.}
    \label{tab:flop_counters}
    \scriptsize
    \begin{tabular}{@{}lrrrrrrr@{}}
        \toprule
        \textbf{Level} & \textbf{N} & \textbf{fv OK} & \textbf{fv 0} & \textbf{fv fail} & \textbf{pt OK} & \textbf{pt 0} & \textbf{pt fail} \\
        \midrule
        L1 & 100 & 57 & 40 & 3 & 61 & 37 & 2 \\
        L2 & 100 & 100 & 0 & 0 & 100 & 0 & 0 \\
        L3 & 50 & 44 & 6 & 0 & 48 & 2 & 0 \\
        L4 & 20 & 1 & 0 & 19 & 19 & 0 & 1 \\
        \midrule
        \textbf{All} & \textbf{270} & \textbf{202} & \textbf{46} & \textbf{22} & \textbf{228} & \textbf{39} & \textbf{3} \\
        & & \multicolumn{3}{c}{\textbf{75\% coverage}} & \multicolumn{3}{c}{\textbf{84\% coverage}} \\
        \bottomrule
    \end{tabular}
\end{minipage}
\hfill
\begin{minipage}[t]{0.45\textwidth}
    \centering
    \caption{\textbf{Zero-shot LLM SOL accuracy on KernelBench.}  Direct prompting achieves 83\% overall but collapses on L3 (38\%) due to spatial dimension tracking failures.}
    \label{tab:llm_zeroshot}
    \scriptsize
    \begin{tabular}{@{}lrrrl@{}}
        \toprule
        \textbf{Level} & \textbf{N} & \textbf{Correct} & \textbf{Acc.} & \textbf{Top error mode} \\
        \midrule
        L1 & 100 & 96 & 96\% & wrong formula \\
        L2 & 100 & 100 & 100\% & --- \\
        L3 & 50 & 19 & 38\% & no spatial tracking \\
        L4 & 20 & 9 & 45\% & wrong formula \\
        \midrule
        \textbf{All} & \textbf{270} & \textbf{224} & \textbf{83\%} & \\
        \bottomrule
    \end{tabular}
\end{minipage}
\end{table}

\textbf{FLOPs and parameter counters.}
Tools such as fvcore~\cite{fvcore} and ptflops~\cite{ptflops} are limited to PyTorch inputs and cover only 75--84\% of operators on KernelBench (\Cref{tab:flop_counters}).
Because they count MACs and parameters but not total memory traffic, they cannot be used to derive precise and tight SOL bounds.

\textbf{Hardware profilers.}
Profilers such as NCU~\cite{ncu2023} and NSys~\cite{nsys2023} measure hardware utilization for a \emph{specific implementation}, i.e., they report how close a kernel gets to peak throughput of each hardware unit.
This answers: \emph{how well does this implementation use the hardware?}
\solar answers a different question: \emph{what is the minimum possible execution time for this algorithm on this hardware?}
The distinction matters: not all algorithms can saturate all hardware units simultaneously.
A memory-bound softmax will never reach 100\% compute utilization; NCU would report headroom that is physically unachievable.
\solar's bound is implementation-agnostic and tighter, considering both algorithmic properties and hardware constraints.

\textbf{Direct LLM-based SOL estimation.}
Zero-shot LLM estimation (Claude Code with Opus~4.6) achieves 83\% accuracy on KernelBench but collapses on composite workloads with 38\% on L3 subgraphs, 45\% on L4 full models and with 20 of 46 errors exceeding $10\times$ (\Cref{tab:llm_zeroshot}).
Moreover, analyses that require systematic search such as cache-aware tiling over buffer capacity constraints~\cite{orojenesis} are difficult to perform reliably through next-token prediction alone.

\textbf{Performance modeling frameworks.}
Timeloop~\cite{parashar2019timeloop} and AccelForge~\cite{accelforge} provide cache-aware performance modeling but require manual workload specification in their input formats; manual translation is error-prone and unvalidated.
\solar bridges this gap by automatically translating source code into a validated einsum representation that these frameworks can consume directly.

\textbf{LLM code translation.}
Frontier LLMs achieve 96\% pass@1 on HumanEval~\cite{chen2021codex}, 80\% on SWE-bench Verified~\cite{jimenez2024swebench}, and 100\% on KernelBench~\cite{ouyang2025kernelbench} with agentic refinement, establishing strong source-to-source translation capabilities.
\solar leverages this capability for verified code translation to enable automatic SOL derivation.

%% file: sections/3_method.tex
\section{Methodology}
\label{sec:overview}
\label{sec:method}

\solar derives SOL bounds from source code through a three-stage pipeline (\Cref{fig:overview}). First, the
\textbf{Validated Agentic Frontend} (\Cref{sec:frontend}) translates the source program into an executable \emph{Affine Loop Intermediate Representation}.
\textbf{Deterministic Einsum Generation} (\Cref{sec:einsum_lifting}) then compiles the validated IR into an \emph{Einsum Graph}, a DAG of extended einsums.
Finally, the \textbf{SOL Analysis} (\Cref{sec:sol_backend}) stage computes multi-fidelity roofline bounds against target hardware specs.

\begin{figure}[t]
\centering
\resizebox{\linewidth}{!}{%
\begin{tikzpicture}[
  node distance=0.48cm and 0.62cm,
  box/.style={draw, rounded corners, align=center, minimum height=0.72cm, font=\scriptsize\sffamily, inner xsep=5pt},
  soft/.style={box, fill=blue!12, minimum width=1.65cm},
  hard/.style={box, fill=green!12, minimum width=1.65cm},
  data/.style={box, fill=orange!12, minimum width=1.45cm},
  check/.style={draw, diamond, aspect=1.8, fill=red!12, align=center, inner sep=1pt, font=\scriptsize\sffamily},
  arrow/.style={-{Stealth[length=2.3mm]}, thick},
  lbl/.style={font=\tiny\sffamily\bfseries, black, align=center}
]
\node[data] (src) {Source\\PyTorch / JAX};
\node[soft, right=of src] (llm) {LLM\\Agent};
\node[data, right=of llm] (dsl) {Affine\\Loop IR};
\node[check, right=of dsl] (val) {output\\match?};
\node[hard, right=of val] (lift) {Deterministic\\Einsum Generation};
\node[data, right=of lift] (graph) {Einsum\\Graph};
\node[hard, right=of graph] (sol) {SOL\\Analysis};
\node[data, below=0.62cm of sol] (out) {SOL Bounds and\\Optimization Hints};
\node[data, above=0.52cm of sol] (arch) {Hardware\\Spec ($\Pi$, $\beta$)};
\draw[arrow] (src) -- (llm);
\draw[arrow] (llm) -- (dsl);
\draw[arrow] (dsl) -- (val);
\draw[arrow] (src.south) |- ++(0,-0.72) -| (val.south);
\draw[arrow] (val) -- node[above,font=\tiny] {pass} (lift);
\draw[arrow] (lift) -- (graph);
\draw[arrow] (graph) -- (sol);
\draw[arrow] (arch) -- (sol);
\draw[arrow] (sol) -- (out);
\draw[arrow, dashed, gray] (val.north) -- ++(0,0.48) -| node[pos=0.27,below,font=\tiny] {retry with mismatch} (llm.north);
\coordinate (shade_top_l) at ($(llm.north)+(0,0.68)$);
\coordinate (shade_top_s) at ($(sol.north)+(0,0.68)$);
\begin{scope}[on background layer]
  \node[draw=green!80, fill=green!4, rounded corners, fit=(llm)(dsl)(val)(shade_top_l), inner sep=4pt, label={[lbl]above:Validated Agentic Frontend}] {};
\end{scope}
\end{tikzpicture}%
}
\caption{\textbf{\solar three-stage pipeline.}
(1)~An agentic front-end translates source code into an executable Affine Loop IR, validated by output comparison.
(2)~A deterministic compiler lifts the validated IR into an Einsum Graph representation. 
(3)~A SOL backend uses the Einsum Graph to compute multi-fidelity roofline bounds against target hardware specs.} 

\label{fig:overview}
\end{figure}

\subsection{Validated Agentic Frontend}
\label{sec:frontend}

This section describes the first stage of the \solar pipeline: a
language-agnostic frontend that uses an LLM agent to translate tensor
programs into an executable intermediate representation called the
\emph{Affine Loop IR}.  The key idea is to treat the LLM as a
programmable, retargetable parser whose output is validated by
numerical comparison against the original program.  LLM code
generation cannot be trusted to produce correct output on the first
attempt, so numerical validation closes this trust gap by providing a
concrete pass/fail signal that gates downstream consumption.  

\subsubsection{Affine Loop IR}
\label{sec:affine-ir}

Using an LLM agent as a front-end for einsum generation places four requirements on
the IR:
\begin{enumerate}[leftmargin=*,itemsep=1pt,topsep=2pt]
  \item \textbf{Einsum extraction.} The representation must enable
        mechanical lifting to einsums by the backend.
  \item \textbf{Expressiveness.} The IR must cover a wide range of DL
        workloads. 
  \item \textbf{LLM generation.} The IR must be easy for an LLM to
        produce correctly.
  \item \textbf{Validation.} The LLM-generated IR must be executable and scalable for numerical validation.
\end{enumerate}

To enable mechanical einsum extraction while retaining expressiveness, 
the Affine Loop IR (\Cref{fig:pipeline_example}b) splits a program into two parts:
restricted affine loop kernels that represent individual einsum operators
and an unrestricted composition layer that handles a wide range of DL
workloads.  

Each \textbf{Affine Loop Kernel} is restricted so that einsum equations can be derived
mechanically from its structure.  A \texttt{@kernel} function
contains one or more perfectly-nested loop nests over
named-dimension tensors.  Loop bounds are static tensor-shape
constants, index expressions are affine in the loop variables, there
is no data-dependent branching, and loop bodies contain only scalar
arithmetic and a small set of fixed builtins.

\textbf{Tensors} carry the metadata the backend needs to derive byte counts.  For example, \texttt{Tensor("B,M,K", dtype="fp16",
role="weight", sparsity=0.5, fill=0)} declares an fp16 weight tensor
with 50\% sparsity and named dimensions \texttt{B}, \texttt{M}, and
\texttt{K}. These named dimensions act
as both loop dimensions and einsum subscripts, so the backend can derive an einsum equation
from the loop nest.

To cover a wide range of DL workloads, an unrestricted top-level
\textbf{Composition Layer} (\texttt{model()}) composes kernels,
invokes library builtins, and expresses recurrence through control flow. 
Builtins handle patterns that do not fit the affine
kernel model: data-movement operations (reshape, slice, stack,
concat), data-dependent indexing (gather), and sparse operations
(\texttt{apply\_mask} which propagates the tensor's \texttt{sparsity}
field so that downstream byte and FLOP estimates reflect effective
sparsity). 


Finally, for ease of LLM generation and fast numerical validation, the IR
is implemented as an \textbf{executable Python library} with Numba support.
LLMs are already strong Python generators, so they can produce IR programs without
fine-tuning, and every program can be run directly and compared
against the original without any additional compilation. For scalable validation, 
a loop whose iterations carry no data dependence
may be annotated \texttt{Dim("B", parallel=True)} by the LLM to enable parallel execution during
validation.

\subsubsection{Agentic IR Translation and Validation }
\label{sec:translation}
\label{sec:validation}

In a traditional compiler, supporting a new source language means
building a dedicated parser, type checker, and lowering pass. 
Instead, \solar uses an LLM agent that translates any source program into the shared Affine Loop IR
through a generate-then-verify loop: the agent proposes a
translation, the pipeline validates it against the original program's
outputs, and on failure provides diagnostic feedback that guides the
agent toward a corrected version (\cite{li2022alphacode,shinn2023reflexion}).

A two-part system prompt separates target-IR knowledge from
source-language conventions for easy portability to new languages.
The \emph{target-IR specification} is fixed and defines the Affine
Loop IR grammar and provides few-shot examples.  The 
\emph{source-language adapter} supplies language-specific
conventions. The pipeline first captures a golden
output by executing the original program, invokes the translation agent, executes the
generated Numba-accelerated IR, compares the outputs
and provides feedback to the agent in a loop.
One limitation of our approach is that it does not provide formal guarantees; incorrect translations can pass if sampled inputs do not expose an output mismatch.



Finally, support for a new source language requires only two additions: (1)~a
model-capture module, and (2)~a source-language adapter for the prompt. The LLM thus becomes a
\emph{verifiable language-agnostic parser}: it reads diverse source
languages, but the downstream pipeline only ever receives
translations that numerically match the original program's outputs. 

\begin{figure}[t]
\centering
\begin{subfigure}[t]{0.48\linewidth}
\begin{lstlisting}[language=Python,basicstyle=\ttfamily\scriptsize,breaklines=true,columns=flexible]
class Model(nn.Module):
    def __init__(self):
        self.fc = nn.Linear(256, 512, bias=True)
        self.D = nn.Parameter(torch.randn(512, 64))

    def forward(self, x):
        y = self.fc(x)
        return torch.matmul(y, self.D)

def get_inputs():
    """Return [batch=2, seq=128, hidden=256]."""
    torch.manual_seed(0)
    return [torch.randn(2, 128, 256)]
\end{lstlisting}
\caption{Source Code.}
\end{subfigure}
\hfill
\begin{subfigure}[t]{0.48\linewidth}
\begin{lstlisting}[language=Python,basicstyle=\ttfamily\tiny,breaklines=true,columns=flexible]
@kernel
def linear(X: Tensor("B,M,K"), W: Tensor("N,K"),
           Bias: Tensor("N")) -> Tensor("B,M,N"):
    Y = Tensor(B=X.shape[0], M=X.shape[1], N=W.shape[0])
    for b in Dim("B", parallel=True):
      for m in Dim("M"):
       for n in Dim("N"): Y[b, m, n] = Bias[n]
    for b in Dim("B", parallel=True):
      for m in Dim("M"):
       for n in Dim("N"):
        for k in Dim("K"): Y[b, m, n] += X[b, m, k] * W[n, k]
    return Y

@kernel
def matmul(Y: Tensor("B,M,N"), D: Tensor("N,P")) -> Tensor("B,M,P"):
    Z = Tensor(B=Y.shape[0], M=Y.shape[1], P=D.shape[1])
    for b in Dim("B", parallel=True):
      for m in Dim("M"):
       for p in Dim("P"):
        for n in Dim("N"): Z[b, m, p] += Y[b, m, n] * D[n, p]
    return Z
\end{lstlisting}
\caption{Affine Loop IR (LLM-translated).}
\end{subfigure}

\vspace{-1.8cm}
\begin{minipage}[t]{0.48\textwidth}
\centering
\resizebox{\linewidth}{!}{%
\begin{tikzpicture}[
  node distance=0.4cm,
  op/.style={draw, rounded corners, minimum width=1.8cm, minimum height=0.55cm,
             font=\tiny\sffamily, fill=blue!10, align=center},
  io/.style={draw, rounded corners, minimum width=1.0cm, minimum height=0.4cm,
             font=\tiny\sffamily, fill=gray!10, align=center},
  im/.style={draw, dashed, rounded corners, minimum width=1.0cm, minimum height=0.4cm,
             font=\tiny\sffamily, fill=orange!8, align=center},
  arrow/.style={->, thick, >=stealth}
]
\node[io] (x) {\emph{X}\\\tiny BMK};
\node[io, below=0.25cm of x] (w) {\emph{W}\\\tiny NK};
\node[op, right=0.6cm of $(x)!0.5!(w)$] (lin) {linear\\\tiny BMK,NK,N$\to$BMN};
\node[op, right=1.4cm of lin] (mm) {matmul\\\tiny BMN,NP$\to$BMP};
\node[im, above=0.3cm of $(lin.east)!0.5!(mm.west)$] (y) {\emph{Y}\\\tiny BMN};
\node[io, above=0.3cm of mm] (d) {\emph{D}\\\tiny NP};
\node[io, right=0.5cm of mm] (z) {\emph{Z}\\\tiny BMP};
\draw[arrow] (x) -- (lin);
\draw[arrow] (w) -- (lin);
\draw[arrow] (lin) -- (y);
\draw[arrow, dashed, black] (y) -- (mm);
\node[font=\tiny, color=black] at ($(lin.east)!0.5!(mm.west)+(0,-0.28)$) {fusible};
\draw[arrow] (d) -- (mm);
\draw[arrow] (mm) -- (z);
\end{tikzpicture}%
}
\vspace{4pt}
{\scriptsize
\textbf{Rank Sizes:} B{=}2, M{=}128, K{=}256, N{=}512, P{=}64}\\[3pt]
{\tiny
\begin{tabular}{@{}llll@{}}
\toprule
\textbf{Node} & \textbf{Einsum} & \textbf{MACs} & \textbf{Memory} \\
\midrule
\texttt{linear\#0} & BMK,NK,N$\to$BMN & BMKN & BMK+NK+N+BMN \\
\texttt{matmul\#0} & BMN,NP$\to$BMP & BMNP & BMN+NP+BMP \\
\bottomrule
\end{tabular}
}
\\[4pt]
{\small (c) Einsum Graph}
\end{minipage}
\hfill
\begin{minipage}[t]{0.48\textwidth}
\centering
\vspace{6pt}
{\scriptsize
\begin{tabular}{@{}lrrrl@{}}
\toprule
\textbf{Mode} & \textbf{Compute} & \textbf{Memory} & \textbf{SOL} & \textbf{Bottleneck} \\
\midrule
Unfused & 83.9\,MFLOP & 2.03\,MB & 1.00\,$\mu$s & Memory \\
Fused & 83.9\,MFLOP & 0.99\,MB & 0.82\,$\mu$s & Compute \\
\bottomrule
\end{tabular}
}
\\[4pt]
{\small (d) SOL Analysis (H100 PCIe)}
\end{minipage}
\caption{\textbf{End-to-end \solar example} (LinearBiasMatmul).  (a)~PyTorch source.  (b)~Agent-translated Affine Loop IR with named-dimension tensors and affine loops.  (c)~Einsum Graph (fusible edge dashed) with extracted einsum equations.  (d)~SOL analysis: fusion eliminates the intermediate, shifting the bottleneck from memory to compute.}
\label{fig:pipeline_example}
\end{figure}

\subsection{Deterministic Einsum Generation}
\label{sec:einsum_lifting}

The deterministic backend lifts the validated Affine Loop IR into an
\emph{Einsum Graph} representation from which all performance quantities are
derived mechanically. 

\subsubsection{Einsum Graph}
\label{sec:einsum_ir}

\solar lifts to einsums rather than analyzing the Affine Loop IR directly
because the einsum subscript structure provides the minimal
canonical form from which all roofline quantities can be derived.

The Einsum Graph is a DAG whose nodes are \emph{extended
einsums}~\cite{odemuyiwa2024edge,kjolstad2017tensor} and whose edges
are tensors that encode data dependencies between operations
(\Cref{fig:pipeline_example}c).  An einsum specifies a tensor
contraction via subscript strings that name the dimensions of each
operand and the output (e.g.\ \texttt{BMK,KN$\to$BMN} for matrix
multiplication, where \texttt{K} is summed over)~\cite{einstein1922general}.  Extended einsums
generalize this to support reduction operators beyond summation,
e.g.\ max-reduction for softmax.  The einsum representation makes
compute cost and memory traffic derivable in closed form from the
subscript structure alone.  Given an einsum with output dimensions
$\mathcal{D}_O$, reduction dimensions $\mathcal{D}_R$, and input
tensors $\{T_i\}$:
\begin{itemize}[leftmargin=*,itemsep=1pt,topsep=2pt]
    \item \emph{MACs} $= \prod_{d \in \mathcal{D}_O \cup \mathcal{D}_R} |d|$\,: one multiply-accumulate per point in the full iteration space.
    \item \emph{Memory traffic} $= \sum_i \texttt{sizeof}(T_i)\!\cdot\!\prod_{d \in \mathcal{D}_i} |d| \;+\; \texttt{sizeof}(T_{\text{out}})\!\cdot\!\prod_{d \in \mathcal{D}_O} |d|$\,: each tensor is read (or written) once at its full size.
\end{itemize}
These formulas are exact for a single unfused einsum and require only
the dimension sizes and dtypes.  Fusion analysis extends this: when
two adjacent einsums share an intermediate tensor, its traffic is
eliminated from the fused cost, and the combined iteration space
determines the new MAC count.

\subsubsection{Affine Loop IR to Einsum Translation}
\label{sec:conversion_flow}

The backend first extracts a dataflow graph from the verified IR by
running the \texttt{model()} function under a light-weight tracer
that intercepts every kernel and builtin call, records its inputs and
outputs, and constructs the graph without executing any loop bodies.
Nodes in the resulting graph represent kernel and builtin
invocations, and edges are the tensors that flow between them, each
carrying its associated metadata.  This graph encodes the full
dependency structure of the program and is the basis for all
subsequent analysis. Next, from each kernel node, the backend reads named dimensions,
identifies reduction dimensions (absent from the output), and emits
einsum strings (\Cref{fig:pipeline_example}c).

\begin{definition}[Einsum extraction]
Given a \texttt{@kernel} with output dimensions $\mathcal{D}_O$ and loop indices $\mathcal{D}$, the reduction dimensions are $\mathcal{D}_R = \mathcal{D} \setminus \mathcal{D}_O$.
Each input tensor $T_i$ accessing dimensions $\mathcal{D}_i$ yields an einsum operand $\mathcal{D}_i \to \mathcal{D}_O$ with implicit summation over $\mathcal{D}_R$.
\end{definition}

The Einsum Graph annotates each node with derived quantities (MACs, memory traffic, arithmetic intensity) and fusibility.
Edges encode data dependencies via intermediate tensors, and the graph structure determines which operators can share on-chip buffers and eliminate DRAM round-trips when fused (\Cref{fig:pipeline_example}c).
Each translated program additionally emits a YAML sidecar that records the graph in language-independent form, enabling downstream tools such as Timeloop~\cite{parashar2019timeloop} and AccelForge~\cite{accelforge} to consume the workload. 
The backend detects non-einsum nodes (builtins such as
\texttt{gather} and \texttt{select}, and 
\texttt{reshape} kernels)
and classifies them as data-movement-only. The SOL analysis assigns these nodes zero MACs while
still deriving their memory traffic from the tensor dimensions.

\subsection{SOL Analysis}
\label{sec:sol_backend}

Given an Einsum Graph and architecture specs (peak dense
throughput $\Pi$ in FLOP/s and DRAM bandwidth $\beta$ in B/s),
\solar computes the roofline lower
bound~\cite{williams2009roofline}
(\Cref{fig:pipeline_example}d):
\begin{equation}
    T_{\text{SOL}} = \max\!\left(\frac{\text{FLOPs}}{\Pi},\; \frac{\text{Traffic}}{\beta}\right)
    \label{eq:roofline}
\end{equation}
A kernel must both execute all FLOPs and transfer all data; assuming perfect overlap, the slower term dominates.
The \emph{arithmetic intensity} $\mathcal{A} = \text{FLOPs}/\text{Traffic}$ determines the bottleneck regime: compute-bound when $\mathcal{A} > \Pi/\beta$, memory-bound otherwise.
The boundary $\Pi/\beta$ is the \emph{ridge point}.

\textbf{Graph-level analysis.}
Unlike per-kernel roofline, \solar operates on the full Einsum Graph.
Fusing adjacent operators eliminates intermediate DRAM round-trips, reducing Traffic and tightening $T_{\text{SOL}}$.
\solar infers fusibility from data dependencies in the graph.
On KernelBench L3, fused-graph SOL reveals $7.8\times$ additional headroom beyond per-operator analysis.

\textbf{Multi-fidelity modes.}
\solar provides three SOL fidelity levels, each refining the traffic estimate in \Cref{eq:roofline}.
All three share the same FLOPs; they differ only in how DRAM traffic is counted.
\begin{itemize}[leftmargin=*,itemsep=1pt,topsep=2pt]
    \item \emph{Unfused}: each operator is analyzed independently; all intermediate tensors are read from and written to DRAM between operators.
    Within each operator, \textbf{unlimited on-chip cache} is assumed---every element is transferred exactly once.
    $T_{\text{SOL}}^{\text{unfused}} = \sum_v T_{\text{SOL}}(v)$.
    \item \emph{Fused}: adjacent fusible operators share on-chip buffers, eliminating intermediate DRAM round-trips.
    The same \textbf{unlimited cache} assumption applies: each remaining input/output element is transferred exactly once.
    \item \emph{Cache-aware fused (\orojenesis)}: models \textbf{finite on-chip buffer capacity} with fusion.
    When the working set of a (fused) operator exceeds the hardware cache, tiling forces elements to be re-read from DRAM.
    \orojenesis~\cite{orojenesis} formulates tiling as constrained optimization over data-movement volumes subject to buffer-capacity constraints, solved via integer linear programming on affine access maps using AccelForge~\cite{accelforge}.
\end{itemize}


\textbf{Inverse roofline.}
Given a target latency $T_{\text{target}}$, \solar derives minimum hardware specs: $\Pi_{\text{min}} = \text{FLOPs}/T_{\text{target}}$ and $\beta_{\text{min}} = \text{Traffic}/T_{\text{target}}$.
This enables hardware architects to evaluate whether a proposed platform meets deployment constraints before silicon exists.

\textbf{Limitation.}
\label{sec:limit}
A current limitation of \solar{} is that its analysis is based solely on tensor shapes rather than values. Consequently, it cannot capture value-dependent optimizations such as compression or constant propagation, and may overlook performance gains from structured or repeated data that enable more efficient memory access or algebraic simplifications. 
Additionally, the SOL bound may not be tight in practice due to hardware variability, such as power capping or thermal throttling.

%% file: sections/4_eval.tex
\section{Evaluation}
\label{sec:eval}

We evaluate \solar on KernelBench~\cite{ouyang2025kernelbench} (270 problems, L1--L4; PyTorch, H100 PCIe), JAX/Flax workloads (8 programs), and robotics models (3 configurations on Jetson Thor), organized around three practitioner questions (\Cref{sec:headroom,sec:algo_opt,sec:hw_provision}).

\subsection{How Much Headroom Exists and How Do I Close It?}
\label{sec:headroom}

\begin{figure}[t]
\centering
\begin{subfigure}[b]{0.48\textwidth}
    \centering
    \includegraphics[width=\textwidth]{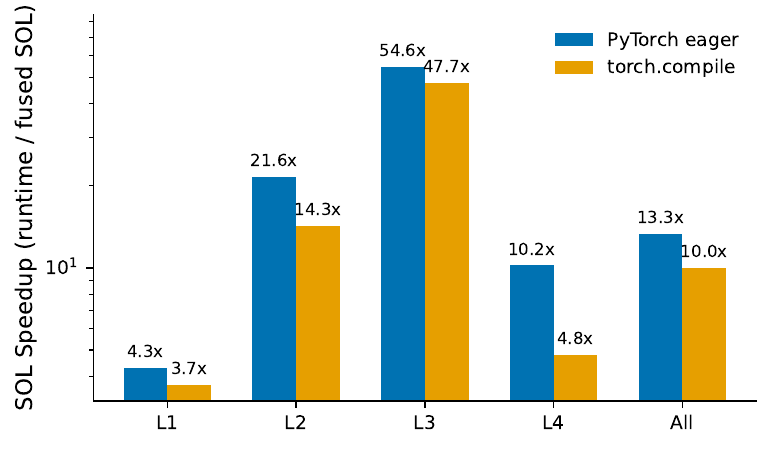}
    \caption{KernelBench headroom (L1--L4).}
    \label{fig:kb_headroom}
\end{subfigure}
\hfill
\begin{subfigure}[b]{0.384\textwidth}
    \centering
    \includegraphics[width=\textwidth]{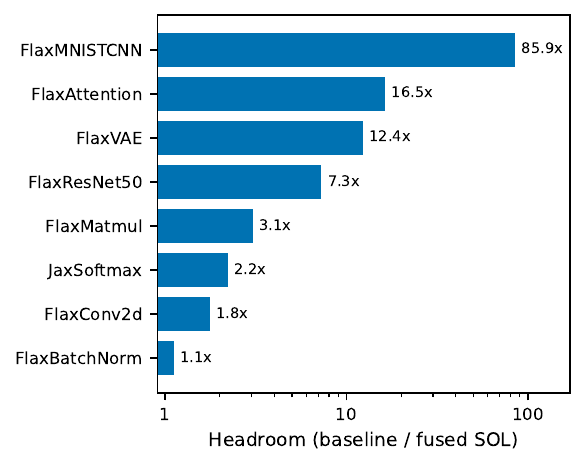}
    \caption{JAX/Flax headroom.}
    \label{fig:jax_headroom}
\end{subfigure}
\caption{\textbf{SOL headroom analysis.}  (a)~KernelBench: measured runtime over fused SOL across L1--L4 levels.  (b)~JAX/Flax: headroom (baseline / fused SOL) spans from $1.1\times$ (BatchNorm) to $85.9\times$ (MNISTCNN), demonstrating language-agnostic frontend coverage.}
\label{fig:headroom_analysis}
\end{figure}

\begin{figure}[t]
\centering
\begin{subfigure}[b]{0.48\textwidth}
    \centering
    \includegraphics[width=\textwidth]{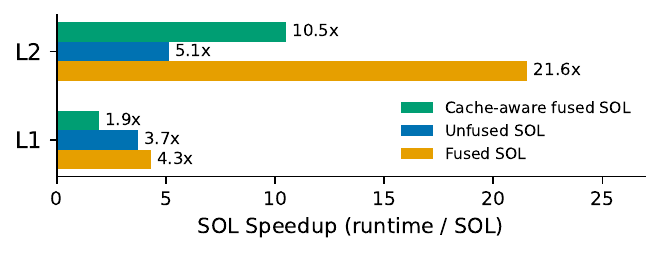}
    \caption{Multi-fidelity SOL speedup for L1/L2.  Cache-aware is tightest on L1; unfused is tightest on L2.}
    \label{fig:fused_unfused}
\end{subfigure}
\hfill
\begin{subfigure}[b]{0.48\textwidth}
    \centering
    \includegraphics[width=\textwidth]{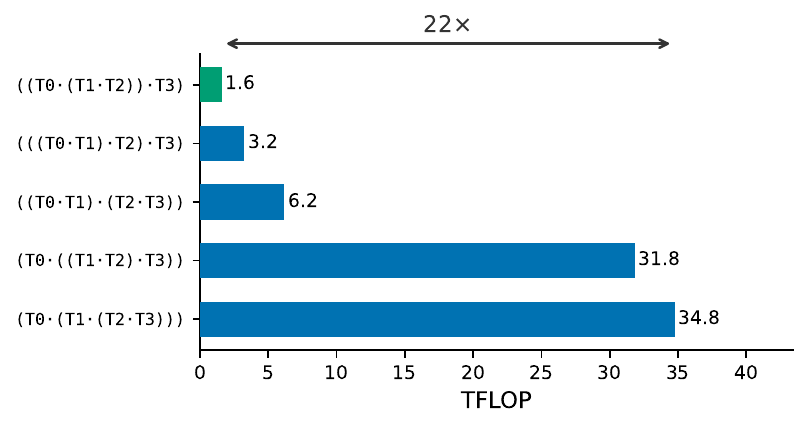}
    \caption{Einsum chain reordering for DeepSeek MLA.}
    \label{fig:mla_reorder}
\end{subfigure}
\caption{\textbf{Optimization hints.}  (a)~Multi-fidelity SOL on L1/L2: on L1, cache-aware is tightest (intra-operator tiling costs dominate); on L2, unfused is tightest (intermediate tensor traffic dominates).  (b)~All 5 association orders of a 4-tensor MLA chain; $22\times$ TFLOP range, with optimal order achieving $2.04\times$ reduction over na\"ive left-to-right.}
\label{fig:opt_hints}
\end{figure}

\textbf{Quantifying the gap.}
\solar achieves 100\% validated analysis coverage on KernelBench.
\Cref{fig:kb_headroom} plots the geomean SOL speedup (runtime / fused SOL) for PyTorch eager and \texttt{torch.compile} across all four levels.
SOL headroom grows with graph complexity: L1 shows $4.3\times$ (eager) and $3.7\times$ (\texttt{compile}), indicating that the compiler captures most of the single-operator gap.
L2 widens to $21.6\times$ vs.\ $14.3\times$---intermediate tensor traffic creates substantial headroom that \texttt{compile} only partially closes.
L3 shows the largest gap ($54.6\times$ eager, $47.7\times$ \texttt{compile}), confirming that graph-level fusion is essential for composite workloads.
On L4, \texttt{compile} closes most of the gap ($4.8\times$ vs.\ $10.2\times$ eager) because it already fuses aggressively on full models.
The gap between eager and \texttt{compile} bars quantifies how much headroom the compiler has already captured; the remaining \texttt{compile} bar height is the headroom that requires kernel-level or algorithmic optimization.
Fused vs.\ unfused SOL analysis (\Cref{fig:fused_unfused}) further decomposes this headroom into fusion and tiling components.
Per-problem breakdowns are in \Cref{app:kb_speedup}; \solbench~\cite{solexecbench2026} (3{,}957 workloads) results are in \Cref{app:solexecbench}.

\textbf{Does it work beyond PyTorch?}
\solar also handles JAX/Flax (\Cref{fig:jax_headroom}).
We evaluated 8 programs spanning single operators to full ResNet-50: headroom ranges from $1.1\times$ (BatchNorm, correctly identifying no fusion benefit) to $85.9\times$ (FlaxMNISTCNN), with FlaxAttention at $16.5\times$.

\textbf{Where should I focus optimization effort?}
Different SOL fidelities surface different bottlenecks (\Cref{fig:fused_unfused}).
On L1, cache-aware SOL ($1.9\times$) is tighter than unfused/fused ($3.7\times$/$4.3\times$) because finite cache forces intra-operator tiling re-reads.
On L2, unfused ($5.1\times$) is tighter than cache-aware ($10.5\times$) because intermediate tensor traffic between isolated kernels dominates.
\orojenesis cache-aware bounds tighten SOL by up to $2.25\times$ overall, identifying cases where tiling strategy---not just fusion---is the bottleneck.
The Einsum graph also enables contraction order enumeration: \Cref{fig:mla_reorder} shows that optimal ordering of DeepSeek MLA's 4-tensor chain achieves $2.04\times$ FLOP reduction over na\"ive left-to-right ($22\times$ range across all 5 orders).

\subsection{How Do I Design Efficient Algorithms for Target Hardware?}
\label{sec:algo_opt}

\solar enables design-space exploration \emph{without hardware access} by sweeping architectural parameters through the analytical pipeline.

\begin{figure}[t]
\centering
\includegraphics[width=\textwidth]{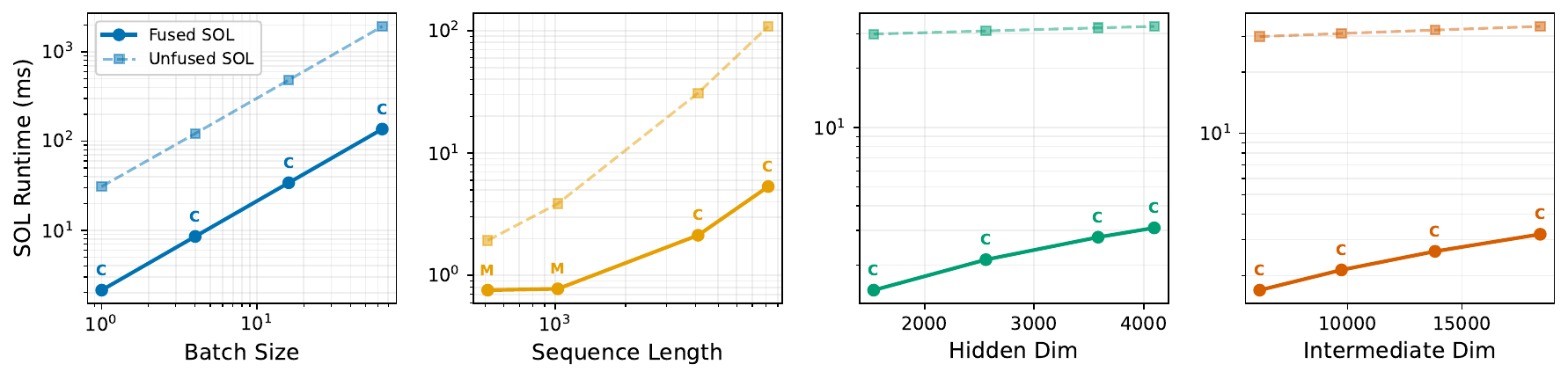}
\caption{\textbf{Qwen3-4B block parameter sensitivity on Jetson Thor.}  Fused and unfused SOL across four architectural axes.  C/M = compute/memory-bound.  Batch size drives linear scaling; sequence length triggers a memory$\to$compute regime shift at ${\sim}$4K tokens.}
\label{fig:qwen3_sweep}
\end{figure}

\textbf{Parameter sensitivity.}
We sweep four axes of a Qwen3-4B block on Jetson Thor (\Cref{fig:qwen3_sweep}): batch size (1--64), sequence length (512--8192), hidden dimension, and intermediate dimension.
Batch size drives linear scaling (2.1\,ms to 136.4\,ms, $64\times$), remaining compute-bound throughout.
Sequence length triggers a memory to compute regime shift: at 512 tokens the block is memory-bound, but at 4K+ tokens quadratic attention traffic pushes it compute-bound.
Hidden and intermediate dimensions show sublinear scaling (${\sim}2\times$) because the MLP dominates and attention traffic is largely dimension-invariant.
Unfused SOL is up to $20\times$ higher than fused (10--15$\times$ typical), confirming fusion is critical across operating points.

\textbf{Cross-platform projection.}
\Cref{fig:cross_hw} projects a Qwen3-4B block onto four platforms: fused SOL spans $5.8\times$ from B200 (0.61\,ms) to A6000 (3.56\,ms).
Fusion benefit ranges from $1.8\times$ on B200 (8\,TB/s) to $14.5\times$ on Jetson Thor (273\,GB/s), quantifying the bandwidth-fusion interaction.

\subsection{What Hardware Do I Need for Real-Time Deployment?}
\label{sec:hw_provision}

\begin{figure}[t]
\begin{minipage}[t]{0.30\textwidth}
    \centering
    \scalebox{1}[1.2]{\includegraphics[width=\textwidth]{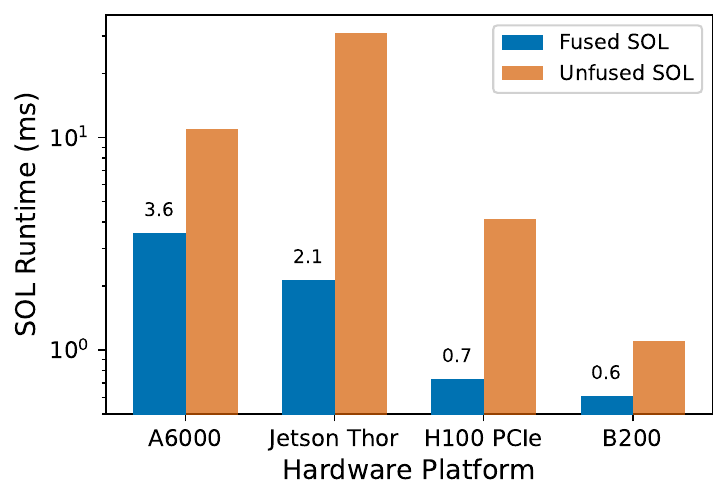}}
    \captionof{figure}{\textbf{Cross-hardware Qwen3-4B block SOL.} \solar enables designers to compare deployment targets and identify the suited platform.}
    \label{fig:cross_hw}
\end{minipage}
\hfill
\begin{minipage}[t]{0.67\textwidth}
    \centering
    \begin{subfigure}[b]{0.48\textwidth}
        \centering
        \includegraphics[width=\textwidth]{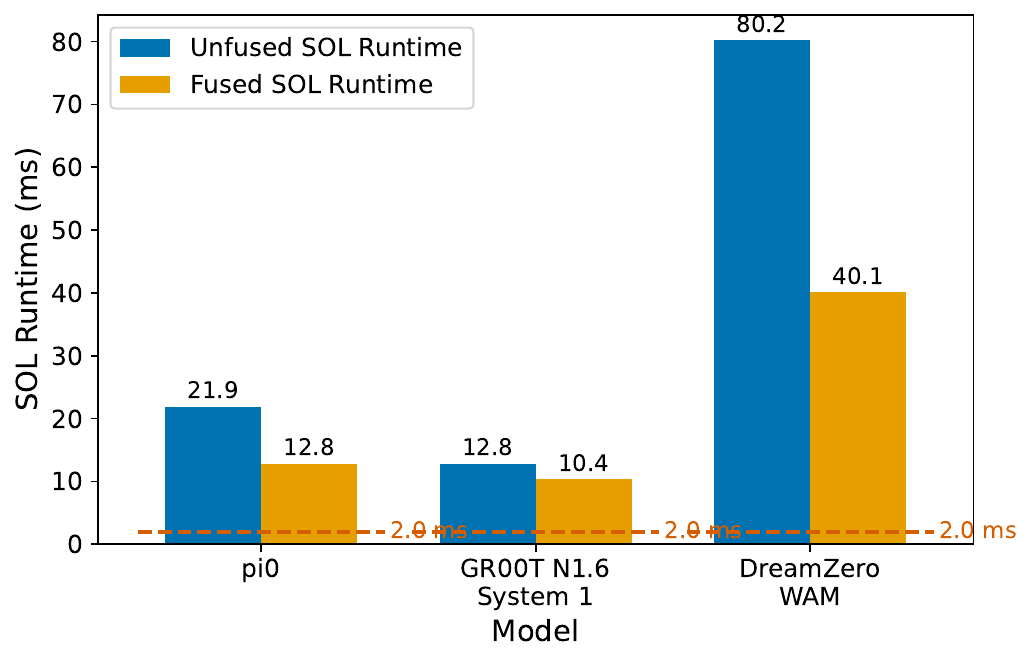}
        \caption{SOL runtime (unfused vs.\ fused).}
        \label{fig:robotics_runtime}
    \end{subfigure}
    \hfill
    \begin{subfigure}[b]{0.48\textwidth}
        \centering
        \includegraphics[width=\textwidth]{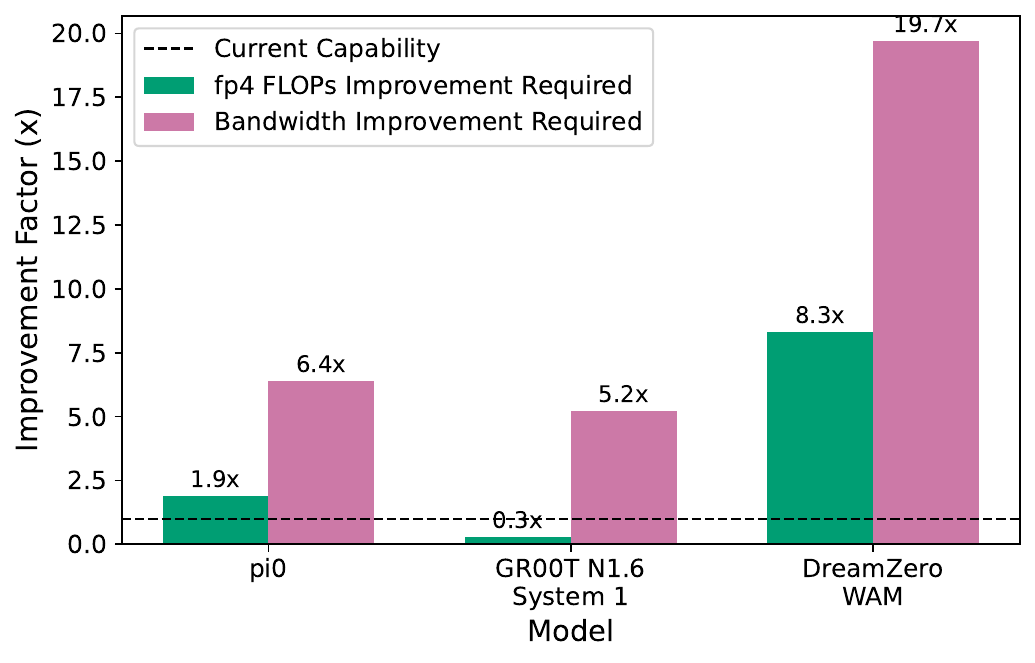}
        \caption{HW improvement required.}
        \label{fig:robotics_hw}
    \end{subfigure}
    \captionof{figure}{\textbf{Robotics model analysis on Jetson Thor.}  (a)~Fused SOL is significantly lower than unfused for all models.  (b)~Bandwidth improvement dominates compute improvement for all models, confirming memory-bound behavior.  DreamZero WAM requires $19.7\times$ current bandwidth for 500\,Hz.}
    \label{fig:robotics}
\end{minipage}
\end{figure}

\solar's inverse roofline derives minimum $\Pi$ and $\beta$ from latency constraints, answering: \emph{what hardware must I provision to meet a deployment target?}
We analyze three robotics models (pi0 (3B), GR00T N1.6 System~1, DreamZero WAM (14B)) on Jetson Thor (1{,}035\,TFLOP/s FP4, 273\,GB/s) targeting 500\,Hz servo control loops (\Cref{fig:robotics}).

All three are memory-bound: pi0 requires $\beta_{\text{min}} = 1{,}750$\,GB/s ($6.4\times$ current) for 500\,Hz; GR00T N1.6 needs only $0.3\times$ current compute, confirming it is entirely bandwidth-limited.
DreamZero WAM requires $19.7\times$ bandwidth \emph{and} $8.3\times$ compute---no single hardware upgrade suffices, pointing to model compression and algorithmic changes as necessary complements to silicon.
Full analysis is in \Cref{app:robotics}.

%% file: sections/5_conclusion.tex
\section{Conclusion}
\label{sec:conclusion}
We presented \solar, the first framework to derive validated
Speed-of-Light bounds directly from PyTorch and JAX source code.
By separating generative translation from deterministic analysis,
\solar's source-to-SOL flow makes SOL derivation accessible without manual modeling or
profiling hardware.  Across KernelBench, JAX/Flax
models, and robotics workloads, \solar
quantifies optimization headroom, identifies fusion and
chain-reordering opportunities, enables cross-platform exploration
in the absence of hardware access, and derives hardware
provisioning targets.





%% file: appendix.tex
\section{KernelBench Details}
\label{app:kernelbench}
\label{app:kb_speedup}

KernelBench~\cite{ouyang2025kernelbench} contains 300 problems organized into four levels of increasing complexity (L1: single operators, L2: operator sequences, L3: model subgraphs, L4: full models).
Our evaluation uses a 270-problem subset (100 L1, 100 L2, 50 L3, 20 L4) for which both SOL and PyTorch baseline runtimes are available.
Geomean SOL speedup (PyTorch eager / fused SOL) ranges from $4.3\times$ on L1 to $54.6\times$ on L3, with L4 at $10.2\times$.

\begin{figure}[H]
\centering
\includegraphics[width=\textwidth]{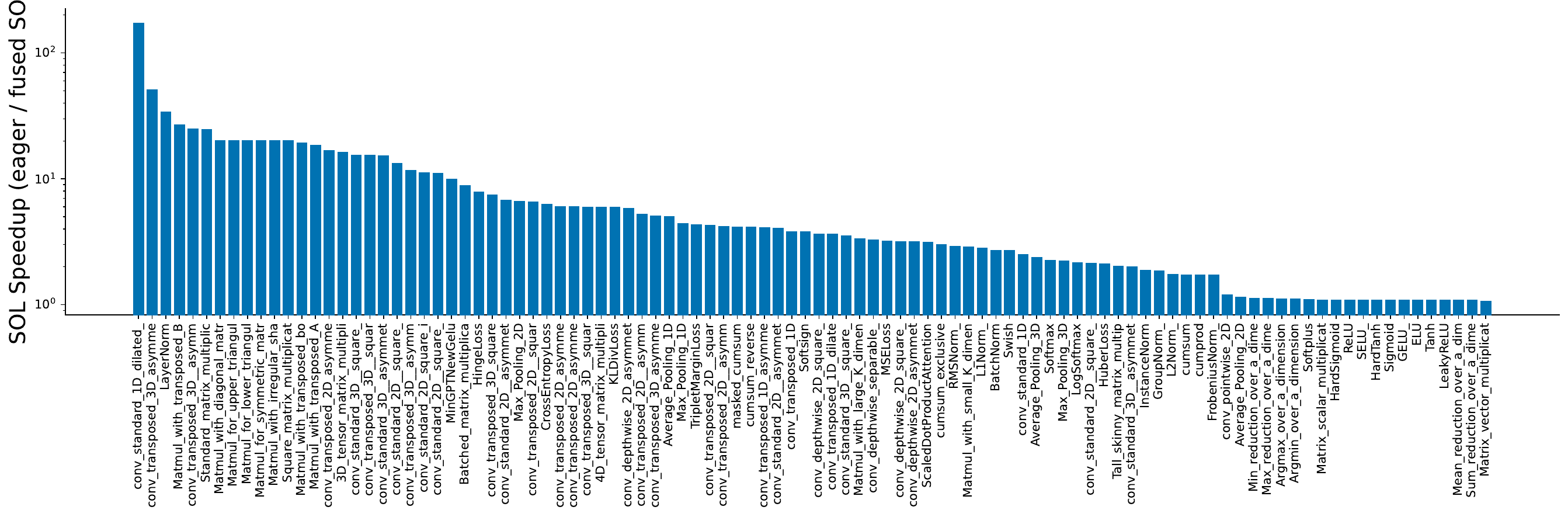}
\caption{KernelBench L1: per-problem SOL speedup (geomean $4.3\times$).}
\label{fig:kb_speedup_l1}
\end{figure}

\begin{figure}[H]
\centering
\includegraphics[width=\textwidth]{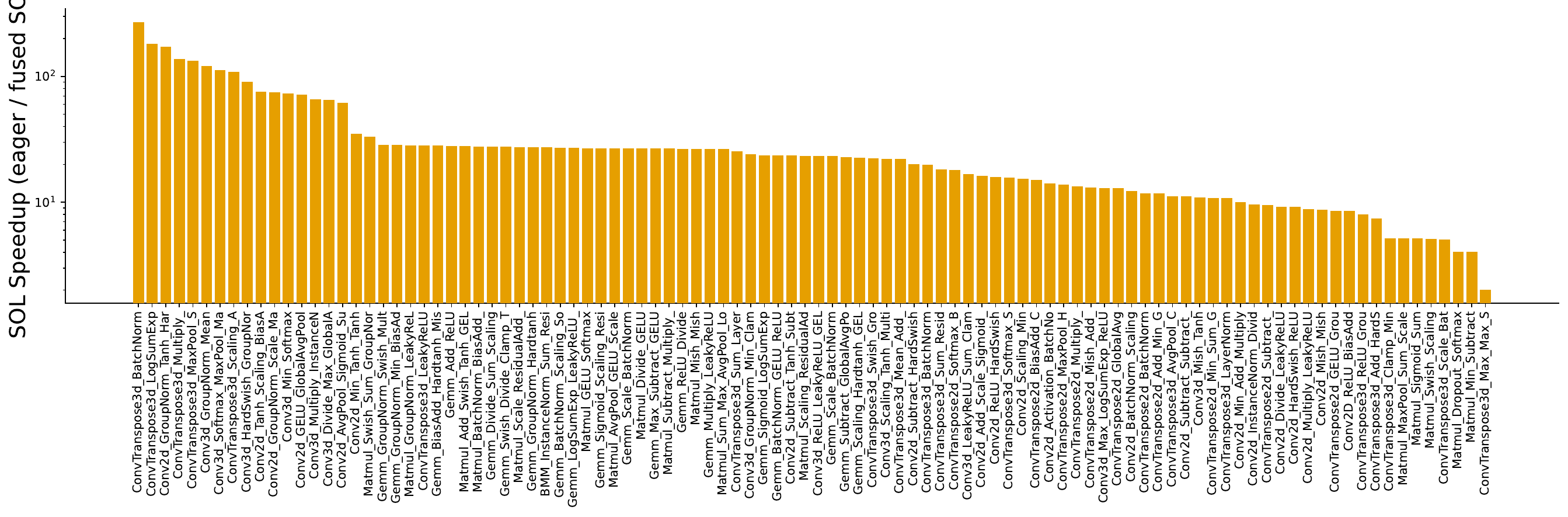}
\caption{KernelBench L2: per-problem SOL speedup (geomean $21.6\times$).}
\label{fig:kb_speedup_l2}
\end{figure}

\begin{figure}[H]
\centering
\includegraphics[width=\textwidth]{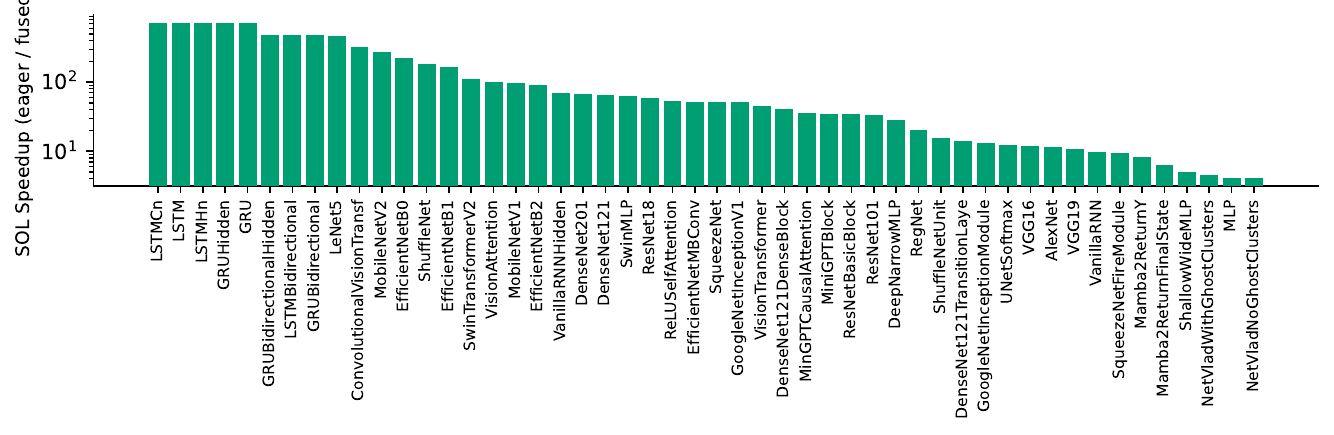}
\caption{KernelBench L3: per-problem SOL speedup (geomean $54.6\times$).}
\label{fig:kb_speedup_l3}
\end{figure}

\begin{figure}[H]
\centering
\includegraphics[width=0.5\textwidth]{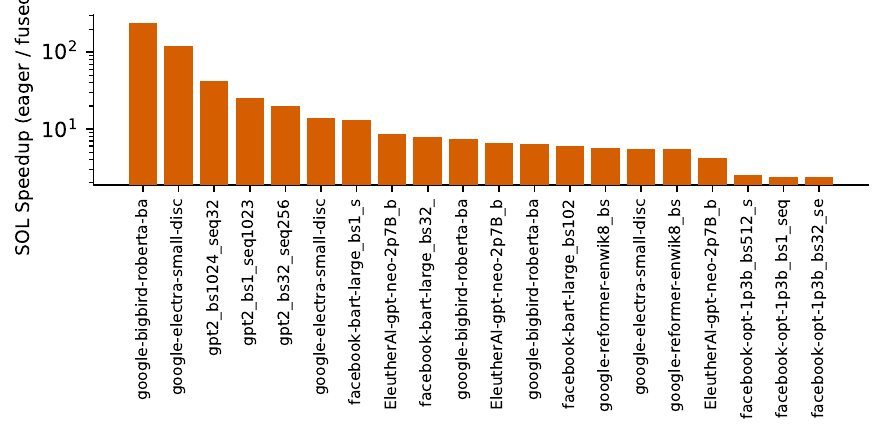}
\caption{KernelBench L4: per-problem SOL speedup (geomean $10.2\times$).}
\label{fig:kb_speedup_l4}
\end{figure}

\Cref{fig:kb_speedup_l1,fig:kb_speedup_l2,fig:kb_speedup_l3,fig:kb_speedup_l4} show per-problem SOL speedup (PyTorch eager runtime / fused SOL) for each level, sorted in descending order.

\section{\solbench SOL Analysis}
\label{app:solexecbench}

\begin{figure}[H]
\centering
\begin{subfigure}[b]{0.48\textwidth}
    \centering
    \includegraphics[width=\textwidth]{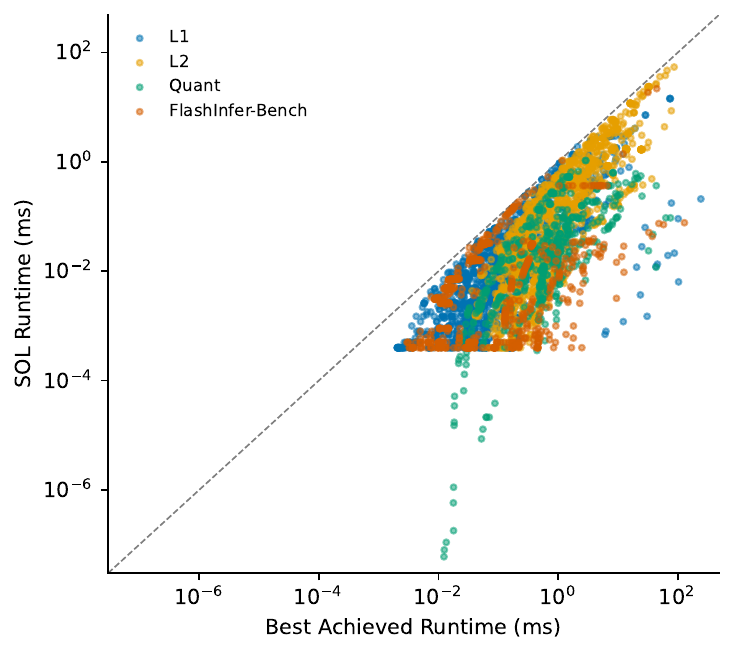}
    \caption{SOL vs.\ best achieved runtime.}
    \label{fig:solexecbench_scatter}
\end{subfigure}
\hfill
\begin{subfigure}[b]{0.40\textwidth}
    \centering
    \includegraphics[width=\textwidth]{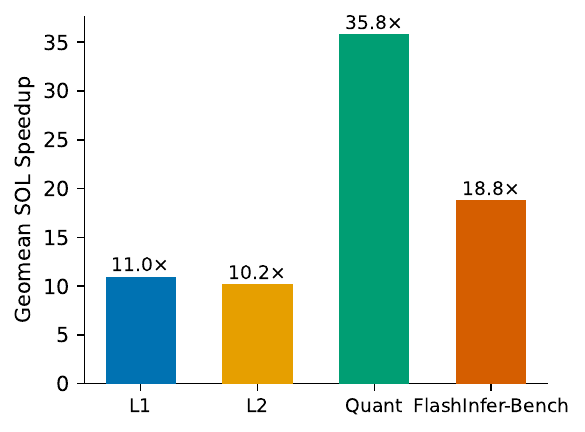}
    \caption{Geomean SOL speedup per subset.}
    \label{fig:solexecbench_geomean}
\end{subfigure}
\caption{\textbf{\solbench SOL analysis.}  (a)~Each point is a workload; gap to the diagonal represents optimization headroom.  (b)~Quant kernels show the largest headroom ($35.8\times$), followed by FlashInfer-Bench ($18.8\times$), L1 ($11.0\times$), and L2 ($10.2\times$).}
\label{fig:solexecbench}
\end{figure}

\solbench~\cite{solexecbench2026} is a benchmark of 3{,}957 real-world GPU kernel workloads spanning four subsets: L1 (1{,}480 single operators), L2 (1{,}299 operator sequences), Quant (518 quantization kernels), and FlashInfer-Bench (660 attention kernels).
Unlike KernelBench, which targets algorithmic diversity, \solbench emphasizes production workload shapes drawn from deployed systems.
We apply \solar to derive SOL scores for all 3{,}957 workloads.
\Cref{fig:solexecbench_scatter} plots SOL runtime against best achieved runtime; points below the diagonal indicate optimization headroom.
\Cref{fig:solexecbench_geomean} summarizes the geometric mean SOL speedup per subset.

\section{Robotics Model SOL Analysis}
\label{app:robotics}

\Cref{tab:robotics_full} provides the full SOL analysis for seven robotics model configurations on Jetson Thor (1{,}035\,TFLOP/s FP4, 273\,GB/s).
All models use W4A4 precision.
The table covers compute and memory characteristics, unfused and fused SOL runtimes, achievable inference frequencies, and the hardware improvements required to meet target deployment frequencies.

\begin{table}[H]
\centering
\caption{\textbf{Full robotics model SOL analysis on Jetson Thor.}  All models are memory-bound.  Target frequencies reflect deployment requirements (500\,Hz for servo control, 30\,Hz for vision).  N/A indicates no target was specified for that configuration.}
\label{tab:robotics_full}
\scriptsize
\begin{adjustbox}{max width=\textwidth}
\begin{tabular}{@{}l|rrrrrrr@{}}
\toprule
\textbf{Metric} & \textbf{pi0} & \textbf{GR00T N1.6} & \textbf{GR00T N1.6 Sys\,1} & \textbf{GR00T N1.6 Sys\,2} & \textbf{DreamZero} & \textbf{DreamZero WAM} & \textbf{DreamZero VLB} \\
\midrule
Precision & W4A4 & W4A4 & W4A4 & W4A4 & W4A4 & W4A4 & W4A4 \\
Denoising Steps & 10 & 4 & 4 & 1 & 4 & 4 & 1 \\
\midrule
Compute (GFLOP) & 4{,}007 & 1{,}725 & 665 & 1{,}059 & 20{,}899 & 17{,}179 & 3{,}720 \\
Data Movement Unfused (GB) & 5.95 & 5.38 & 3.43 & 1.95 & 27.0 & 22.0 & 5.0 \\
Data Movement Fused (GB) & 3.50 & 3.73 & 2.85 & 0.88 & 14.0 & 10.75 & 3.25 \\
Arithmetic Intensity & 1{,}145 & 463 & 233 & 1{,}210 & 1{,}493 & 1{,}598 & 1{,}145 \\
Bottleneck & memory & memory & memory & memory & memory & memory & memory \\
\midrule
Runtime Unfused (ms) & 21.9 & 19.8 & 12.8 & 7.2 & 99.4 & 80.2 & 19.2 \\
Runtime Fused (ms) & 12.8 & 13.6 & 10.4 & 3.2 & 52.0 & 40.1 & 11.9 \\
Frequency Unfused (Hz) & 45.7 & 50.6 & 78.1 & 138.9 & 10.1 & 12.5 & 52.0 \\
Frequency Fused (Hz) & 78.0 & 73.5 & 96.2 & 312.5 & 19.2 & 25.0 & 84.0 \\
\midrule
Target Frequency (Hz) & 500 & N/A & 500 & 30 & N/A & 500 & 30 \\
Speedup Required & $6.4\times$ & N/A & $5.2\times$ & $0.1\times$ & N/A & $20.0\times$ & $0.4\times$ \\
Desired $\Pi$ (TFLOP/s) & 2{,}004 & N/A & 333 & 32 & N/A & 8{,}590 & 112 \\
$\Pi$ Improvement & $1.9\times$ & N/A & $0.3\times$ & $0.03\times$ & N/A & $8.3\times$ & $0.1\times$ \\
Desired $\beta$ (GB/s) & 1{,}750 & N/A & 1{,}425 & 26 & N/A & 5{,}375 & 98 \\
$\beta$ Improvement & $6.4\times$ & N/A & $5.2\times$ & $0.1\times$ & N/A & $19.7\times$ & $0.4\times$ \\
\bottomrule
\end{tabular}
\end{adjustbox}
\end{table}